\pdfoutput=1

\documentclass[11pt]{article}

\usepackage{acl}

\usepackage{times}
\usepackage{latexsym}

\usepackage[T1]{fontenc}

\usepackage[utf8]{inputenc}

\usepackage{microtype}
\usepackage{graphicx}
\usepackage{hyperref}
\usepackage{amsmath}
\usepackage{pifont}
\usepackage{subcaption,booktabs}
\usepackage{natbib}
\usepackage{caption}
\usepackage{multirow}
\usepackage{subcaption}
\usepackage{graphicx}
\usepackage{makecell}
\usepackage{pgfplots}
\usepackage{todonotes}
\usepackage{tablefootnote}
\usepackage{listings}
\usepackage{footnote}

\usepackage[parfill]{parskip}

\DeclareMathOperator*{\desc}{desc}

\newcolumntype{H}{>{\setbox0=\hbox\bgroup}c<{\egroup}@{}}
  
\newcommand{\model}{\textsc{AnyTOD}}
\newcommand{\modelshort}{\textsc{AT}}
\newcommand{\startwo}{\textsc{STARv2}}
\newcommand{\starone}{\textsc{STAR}}

\definecolor{codegreen}{rgb}{0,0.6,0}
\definecolor{codegray}{rgb}{0.5,0.5,0.5}
\definecolor{codepurple}{rgb}{0.58,0,0.82}
\definecolor{backcolour}{rgb}{0.95,0.95,0.92}

\lstdefinestyle{mystyle}{
    backgroundcolor=\color{backcolour},   
    commentstyle=\color{codegreen},
    keywordstyle=\color{magenta},
    numberstyle=\tiny\color{codegray},
    stringstyle=\color{codepurple},
    basicstyle=\ttfamily\footnotesize,
    breakatwhitespace=false,         
    breaklines=true,                 
    captionpos=b,                    
    keepspaces=true,                 
    numbers=left,                    
    numbersep=5pt,                  
    showspaces=false,                
    showstringspaces=false,
    showtabs=false,                  
    tabsize=2
}

\title{\model: A Programmable Task-Oriented Dialog System}

\author{\textbf{Jeffrey Zhao, Yuan Cao, Raghav Gupta, Harrison Lee, Abhinav Rastogi, } \\
\textbf{Mingqiu Wang, Hagen Soltau, Izhak Shafran, Yonghui Wu} \\
Google Research \\
\texttt{\{jeffreyzhao, yuancao\}@google.com}}

\begin{document}
\maketitle

\begin{abstract}
We propose \model, an end-to-end, zero-shot task-oriented dialog (TOD) system capable of handling unseen tasks without task-specific training. We view TOD as a program executed by a language model (LM), where program logic and ontology is provided by a designer as a schema.
To enable generalization to unseen schemas and programs without prior training, \model{} adopts a neuro-symbolic approach. A neural LM keeps track of events occurring during a conversation and a symbolic program implementing the dialog policy is executed to recommend next actions \model{} should take.
This approach drastically reduces data annotation and model training requirements, addressing the enduring challenge of rapidly adapting a TOD system to unseen tasks and domains. We demonstrate state-of-the-art results on \starone{}~\citep{mehri-eskenazi-2021-schema-paradigm},  ABCD~\citep{chen-etal-2021-action} and SGD~\citep{rastogi2020-sgd} benchmarks. We also demonstrate strong zero-shot transfer ability in low-resource settings, such as zero-shot on MultiWOZ~\citep{budzianowski2018large}. In addition, we release \startwo{}, an updated version of the \starone{} dataset with richer annotations, for benchmarking zero-shot end-to-end TOD models.\footnote{The \startwo{} dataset will be released soon.}
\end{abstract}

\section{Introduction} \label{sec:intro}

\begin{figure}[t!]
\includegraphics[width=\columnwidth]{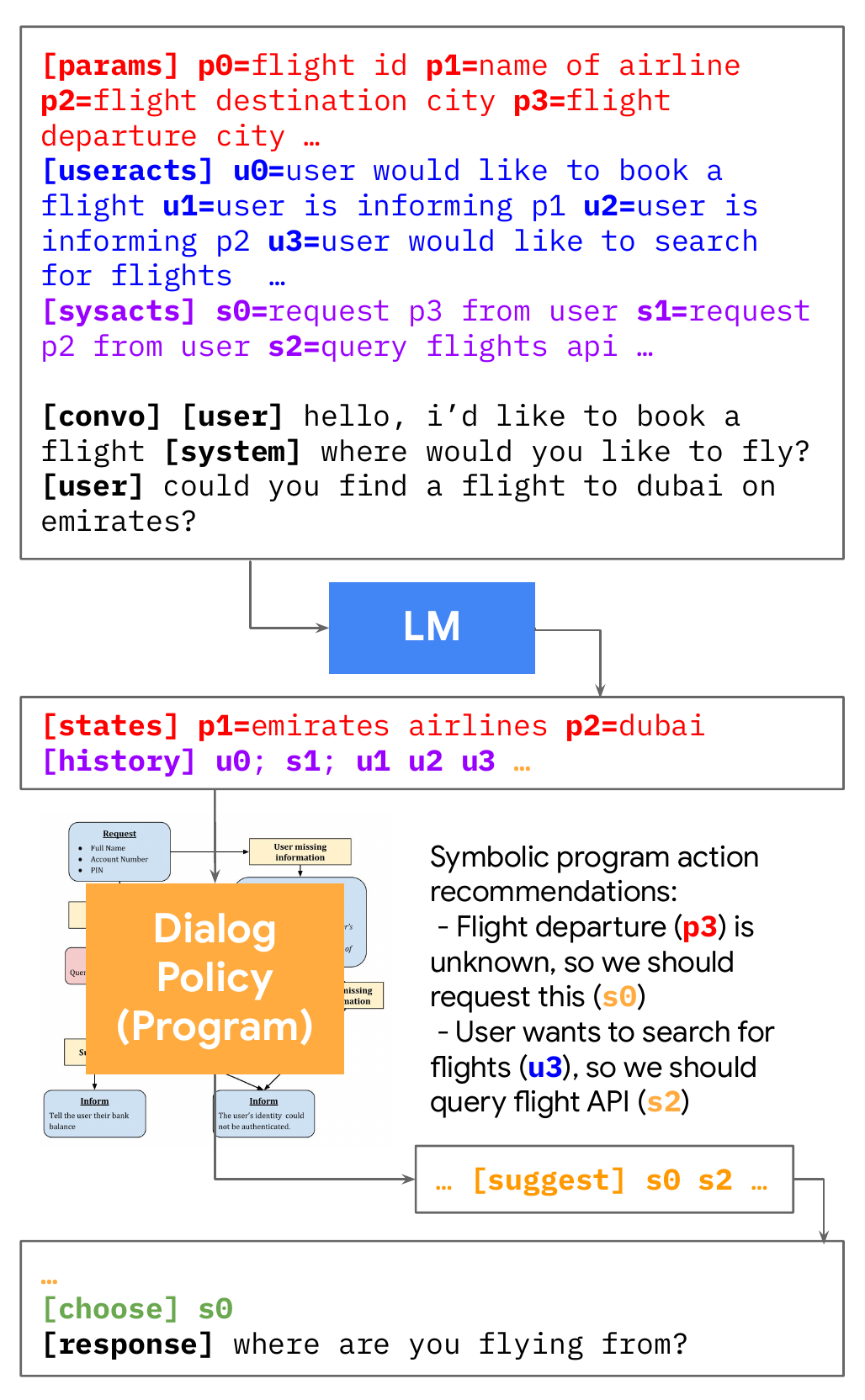}
\centering
\caption{An overview of the \model{} system. 
A LM conducts zero-shot state and action tracking with respect to a provided schema, abstracting it into a sequence of symbols. A program that executes the dialog policy then recommends which actions to take based on the states sequence, the LM then chooses a single final action and generating a response.
}
\label{fig:anytod}
\vspace{-10pt}
\end{figure}

An enduring challenge in building and maintaining task-oriented dialog (TOD) systems is efficiently adapting to a new task or domain. For instance, if we were to add the ability to book flight tickets to an existing system that can only handle booking train tickets, this requires new conversations about flight booking to be manually collected and labelled, as well as retraining of natural language understanding (NLU) and policy models. These data efficiency and scaling problems compound for multi-task TOD systems, as each task may have its own bespoke ontology and policy.

To tackle this problem, we propose \model{}, an end-to-end TOD system that can be \emph{programmed} to support unseen tasks or domains without prior training, significantly speeding up the TOD design process by easing data collection and training requirements. To the best of our knowledge, \model{} is the first end-to-end TOD system capable of zero-shot transfer. To this end, we view TOD as a \emph{program} that a language model (LM) must execute throughout a conversation, and can rely on to provide guidance. Any predefined task policy, implemented as a program, can be used to control \model{}, allowing arbitrary business logic to be executed for a specific task. To demonstrate the efficacy of this paradigm, we experiment with the \starone{} \citep{mehri-eskenazi-2021-schema-paradigm}, ABCD \citep{chen-etal-2021-action}, SGD \citep{rastogi2020-sgd} and MultiWoZ \citep{Eric2019-multiwoz-21} benchmarks. Not only does \model{} achieve state-of-the-art results in full-shot settings, it also achieves high accuracy in zero-shot setups.

\noindent \textbf{Overview of \model{}} To adhere to a given program, \model{} adopts a neuro-symbolic approach (Figure \ref{fig:anytod}). A neural LM is trained for zero-shot dialog state tracking (DST) and action state tracking (AST), abstracting both states and actions into a sequence of symbols. To support zero-shot, we follow the schema-guided paradigm advocated by \citet{rastogi2020-sgd}, and provide a \emph{schema} to the LM as contextual information, describing all parameters and actions that should be tracked in natural language. By training on a large corpus of diverse schemas, the LM generalizes to arbitrary and unseen schemas \citep{Lee2021-dst-with-lm-schema, zhao2022-d3st}. A schema should also provide a symbolic program that declares the task logic, which is executed to \emph{recommend} possible next actions the agent can take, conditioned on the current dialog states. These recommendations are then reincorporated into the LM, which selects a single next action prediction (NAP), and generates a response. Note that the symbolic program forces \model{} to consider a dialog policy explicitly, driving zero-shot transfer onto unseen policies and allowing arbitrarily complex business logic to be employed. However, the program's recommendations are only considered as guidelines, and it is up to the LM to make a final decision on the NAP.

\noindent \textbf{\startwo{}} We also introduce \startwo{}, an improved version of the STAR dataset \citep{mosig-2020-star-dataset}. The original STAR dataset is very valuable for benchmarking zero-shot dialog policy and NAP across a diverse set of tasks or domains, through following a provided \emph{policy graph} that outlines the intended flow of a conversation. However, the original dataset made following these policy graphs difficult, due to its lack of training data for DST and AST. Moreover, we found that the schema entity descriptions provided by the original dataset were not intuitive enough to truly support zero-shot DST and AST. To resolve these limitations, the \startwo{} dataset provides new belief state and action state annotations to the STAR dataset, as well as more intuitive natural language descriptions for many schema elements. In Section \ref{sec:starv2_results}, we show that these changes facilitate stronger zero-shot DST and AST. However, the ground truth NAP on each system turn is left untouched, allowing direct comparison to results trained on the original STAR dataset. We hope that \startwo{} can serve as a new benchmark for TOD systems and drive further research for zero-shot TOD.

\section{Related Work}

\noindent \textbf{Zero-shot Task-oriented Dialog} Fueled by the difficulty of adapting existing TOD systems to new tasks/domains, zero-shot TOD systems have recently seen increasing interest. Much of this work has been on zero-shot DST, with the primary approach being characterizing parameters through names \citep{Wu2019-trade} or descriptions \citep{Lin2021-slot-descs-for-zero-shot, Lee2021-dst-with-lm-schema, zhao2022-d3st}. Another approach has been through in-context finetuning \citep{Shah2019-robust-slot-filling, gupta-etal-2022-show}, in which a labeled exemplar conversation is given as a prompt to a LM. \citet{Mi2021-ks} demonstrated a more comprehensive approach, including task instructions, constraints, and prompts. In general, these results follow the schema-guided paradigm advocated by \citet{rastogi2020-sgd, mosig-2020-star-dataset}.

By contrast, there are fewer results on zero-shot dialog policy (AST and NAP). To the best of our knowledge, the only result is SAM \citep{mehri-eskenazi-2021-schema-paradigm}, which aligns an LM for an unseen dialog policy by following an explicit policy graph. While similar to the policy graph execution we demonstrate in \model{}, there are two differences. First, SAM lacks supervised training on DST and AST, and relies on ground truth NAP only, forcing user state and action tracking to be inextricably linked with the final system action prediction, hurting its ability to generalize to arbitrary policy graphs. Second, SAM is a classification model limited to NAP, and unlike \model{}, cannot support DST or natural language generation (NLG). Indeed, we show that \model{} is empirically more powerful than SAM in Section \ref{sec:starv2_results}.

To the best of our knowledge, no method has yet to combine zero-shot DST, AST, and NAP into an end-to-end TOD system. All existing end-to-end TOD systems \citep{hosseiniasl2020simple, He2021-galaxy, Yang2020-ubar, Peng2020-soloist} are trained and evaluated on the popular MultiWOZ dataset \citep{Eric2019-multiwoz-21}. As a result, these systems are only aware of the policy for MultiWOZ, and are not robust to arbitrary/unseen policies. In contrast, AnyTOD can generalize to arbitrary policies, and we demonstrate strong performance on MultiWOZ without prior training (Section \ref{sec:mwoz_0shot}).

\noindent \textbf{TOD as Programming} Historically, most TOD approaches use an explicit plan-based dialog policy module \citep{Rich1998-collagen, Ferguson1998-trips, Bohus2009-ravenclaw}.
However, the NLU models powering these TOD systems are tightly coupled to a specific plan, and must be retrained for even slight changes to the plan. In contrast, \model{} enables zero-shot dialog policy by training NLU models to be robust to arbitrary programs as policies. Further, \model{} uses the program as contextual information to NLU, and refines its NAP with respect to the conversation, belief state, and action history instead of simply accepting the plan's dictated next action(s).

Recent work has also focused on discovering structure within conversations i.e. a latent schema, policy graph, or program \citep{shi2018-unsupervised-dialog-structure, Yu2022-unsupservised-slot-schema-induction, Xu2020-discover-dialog-structure}. Notably, SMCalFlow \citep{Machines2020-smcalflow} constructs ``dataflow graphs'' from a conversation, parsing semantic intents into executable programs. \citet{Cheng2020-treedst, Shin2021-constrained-lm-few-shot-semantic-parser} further explore this setup. However, these aim to manipulate an external API/database instead of controlling the agent's behavior.

Beyond the scope of TOD, there has been some work in general neuro-symbolic programming with LMs, in which an LM is influenced by the results of a symbolic system. \citet{Nye2021-neurosymb-reasoning} demonstrated a symbolic reasnoning module that accepts or rejects the logical consistency of generations from a neural LM. \citet{Lu2020-neurologic} explored using predicated logic constraints to control lexical aspects from the generation of an LM. However, \model{} is the first application of such an approach to a practical TOD setting.

\section{Methodology} \label{sec:method}

\subsection{The \model{} System} \label{sec:anytod_method}

An overview of the \model{} system is presented in Fig. \ref{fig:anytod}. %
We decompose \model{} into three steps, and describe each step in detail below:
\begin{enumerate}
    \item Schema and program construction: A designer constructs a schema for \model{} to characterize the ontology of a specific task, as well as a policy graph that declares the task logic. %
    \item Zero-shot DST and AST: A LM performs zero-shot DST and AST with reference to the schema, without task-specific training.
    \item Program execution and NAP: The predicted states and action history are passed to the schema program, which upon execution recommends preferred system actions to the agent. These actions are resent to the LM, which predicts the final system action(s) conditioned on these recommendations, conversation history, and belief states.
\end{enumerate}

\noindent \textbf{Schema Construction} The designer is required to construct a schema defining a task's ontology, and provide a program describing business logic. This is the \emph{only} thing \model{} requires from the designer. For example suppose the designer is creating a flight booking chatbot, they must define the parameters to be tracked (e.g. ``flight id'', ``name of the airline''), and enumerate possible actions the user and agent can take (``user saying they would like to search for flights'', ``agent should query flight booking api''). Following the schema-guided paradigm advocated in \citet{rastogi2020-sgd}, each element in this schema is characterized by a short natural language description, allowing the LM to understand its meaning and facilitate zero-shot transfer.
The schema program can be considered as a function that takes in predicted belief states and actions, and dictate possible NAPs following explicit symbolic rules. Examples can be seen in Section~\ref{sec:anytod_programs_appendix}. In general, this program should infer agent actions in response to user behavior (e.g. ``if user wants to search for flights, query the flight search api'').

\noindent \textbf{Zero-shot DST and AST} Adaptation to novel tasks without training data critically hinges on an LM performing zero-shot DST and AST. For this purpose, we adopt and extend the D3ST approach \citep{zhao2022-d3st} due to its flexibility in zero-shot state and action tracking. Specifically, D3ST conducts zero-shot DST in the following way.
Let $p_0, ... p_n$ be the parameters defined in the schema, and let $\desc(p_i)$ denote a parameter's natural language description. Then, construct a parameter context string
\[
\textbf{\texttt{[params] p$0$=$\desc(p_0)$ ... p$n$=$\desc(p_n)$}} \\
\]
Note that the strings \textbf{\texttt{p$0$, ..., p$n$}} are used as indices. Similar context strings are generated for actions for AST. These context strings are concatenated with the entire conversation history, forming the input to the LM. This input is contextualized by the schema information, allowing the LM to refer to the schema, and enabling zero-shot transfer. The target string contains the conversation belief state and history of actions at each turn of the conversation, both in a parseable format. Let $p_{i_0}, \ldots, p_{i_m}$ be the active parameters in the conversation, with corresponding values $v_{i_0}, \ldots, v_{i_m}$. The belief state is represented as
\[
\textbf{\texttt{[state] p$i_0$=$v_{i_0}$;$...$; p$i_m$=$v_{i_m}$}} \\
\]
Note that inactive slots do not appear in the belief state string. In \model{} D3ST is naturally extended to perform zero-shot AST. Note that D3ST's original formulation in \citet{zhao2022-d3st} was limited to DST, but, in principle, D3ST supports tracking arbitrary events that occur during a conversation, as long as their descriptions are provided.
For AST, we build an target string consisting of a history of actions that were active at each turn of the conversation. Let $\textbf{\texttt{u$j$}}$ and $\textbf{\texttt{s$k$}}$ be the format of D3ST indices for user and system actions. Then, an action history string may look like
\[
\textbf{\texttt{[history] u0 u9; s2; u1; s3; ...}}
\]
This denotes that, on the first turn, the user was performing user actions $\textbf{\texttt{u0}}$ and $\textbf{\texttt{u9}}$. On the second turn, the system was performing system action $\textbf{\texttt{s2}}$, and so on. Note that the active actions for each turn are separated by a \textbf{;} character.

\noindent \textbf{Program Execution} The LM's predicted belief states and action history are then parsed and passed to the schema program. This program should execute the dialog policy and control \model{}, by recommending possible NAPs. Section \ref{sec:anytod_programs_appendix} showcases some example programs for \startwo{} tasks. In the example shown in Figure \ref{fig:anytod}, the current conversation state (``user would like to search for flights to Dubai with Emirates'') satisfies multiple dependency rules (``since the user would like to search for flights, query the flight search api'' and ``since the user has not provided their flight departure location, ask the user for it''). These system actions are then passed back to the LM as a string of system action indices.
\[
\textbf{\texttt{[recommend] s0 s2}}
\]
Finally, given the policy graph's recommended actions as extra conditional information, the LM makes predictions about NAP with respect to the conversation, previously predicted belief states and action history. A response is also generated following the action prediction.
\[
\textbf{\texttt{[selected] s2 [response] hello!}}
\]
Note that the selected action need not be one of the actions recommended from the policy graph output, because actual conversations may not rigorously follow the predefined business logic, and violations are common.
This step allows \model{} to ``softly'' execute the policy graph, balancing between the model's belief before and after receiving recommendations.

\noindent \textbf{Zero-shot transfer} \model{}'s zero-shot transfer ability is enabled by a combination of two design considerations. The first is the LM's description-driven state and event tracking. Since this schema information is provided as context, if this LM is trained on a corpus of diverse schemas, it learns to make predictions by ``reading'' and understanding the schema descriptions. This leads to robustness on \model{}'s state and event tracking for unseen schemas, as shown in \citet{zhao2022-d3st}. Moreover, \model{} facilitates zero-shot policy transfer by executing the provided policy graphs as explicit rules, and by similarly training the LM with a large number of policy graphs when selecting a recommended system action.

\subsection{The \startwo{} Dataset} \label{sec:starv2_method}

To train \model{}, we construct \startwo{}, an updated version of STAR with new ground truth belief state and action annotations, supporting supervised training on DST and AST. These annotations were generated from few-shot training with D3ST \citep{zhao2022-d3st}. We first train D3ST on the SGD dataset, then continue finetuning on a few hand-labeled conversations from STAR.\footnote{4 conversations were labeled from each task in \starone{}.} While not the focus of this paper, the labeling of \startwo{} demonstrates the use of few-shot D3ST in labeling unlabeled conversations on new tasks/domains. %

Further, \startwo{} adds new natural language descriptions for actions in \starone{} schemas. Prior work on \starone{} \citep{mosig-2020-star-dataset, mehri-eskenazi-2021-schema-paradigm} leverages template utterances as schema descriptions, which we qualitatively found to not fully outline the complexity of actions e.g., the action \texttt{user\_weather\_inform\_city} has a template utterance of just \texttt{[CITY]}. \startwo{} provides \texttt{user is informing city} as a more natural action description. We show 
in Section \ref{sec:starv2_results} that these actions improve zero-shot AST.

\section{Experiments}

\subsection{Setup}

\noindent \textbf{Datasets} We demonstrate \model{}'s power in zero-shot settings on the following datasets: 

STAR and STARv2: As described in Section~\ref{sec:starv2_method}, we upgrade the original STAR~\citep{mehri-eskenazi-2021-schema-paradigm} dataset to STARv2. The dataset has 24 tasks across 13 domains, many tasks requiring the model to adhere to a novel policy, providing an important zero-shot AST and NAP benchmark.

ABCD~\cite{chen-etal-2021-action}: The design of the ABCD dataset follows a realistic setup, in which an agent's actions must be balanced between the customer's expressed desires and the constraints set by task policies. It is thus a natural fit for the AnyTOD framework for both training and evaluation.

SGD~\cite{rastogi2020-sgd}: SGD is another schema-guided dataset in which schema elements are provided with natural language descriptions to facilitate task transfer. It contains 45 domains and was generated via simulation. Thus, the agent actions and responses follow pre-defined task logic.

MultiWOZ~\cite{budzianowski-mwoz}: MultiWOZ is the standard dataset for benchmarking TOD models. It contains 7 domains and was generated through Wizard-of-Oz \citep{Kelley1984-woz} data collection, leading to natural conversations.

\noindent \textbf{Training} Our implementation is based upon the open-source T5X codebase \cite{roberts2022t5x} initialized with the public T5 1.1 checkpoints\footnote{\url{https://github.com/google-research/text-to-text-transfer-transformer}} as the LM backend. We update the LM code to execute a schema program and reincorporate the results before making the final NAP, as described in Section~\ref{sec:anytod_method}. We experimented on two T5 sizes: base (250M parameters, trained on 16 TPUv3 chips~\citep{jouppi2017in}) and XXL (11B parameters, trained on 64 TPUv3 chips). We otherwise adopt the default T5X finetuning hyper-parameter settings throughout our experiments.

\subsection{Results on \starone{}}
\label{sec:starv2_results}

Table \ref{table:star_results} shows \model{} results on the \startwo{} dataset on the full-shot and zero-shot domain transfer settings, with both ``happy'' and ``unhappy'' conversations.
In full-shot, models train on 80\% of conversations across all tasks, and evaluate on the remaining 20\%. %
The zero-shot domain setting is a leave-one-out cross-validation across the \startwo{} dataset's 13 domains, evaluating quality on an unseen schema in a completely novel domain. The following metrics are used in our report: joint goal accuracy (JGA) to measure DST, user action F1 (UaF1) to measure AST, system action F1 (SaF1) to measure NAP, and response BLEU.\footnote{See Section~\ref{sec:appendix_starv2_metrics_calc} for details on calculating these metrics.}

Each STAR task schema defines the intended dialog policy by providing a \emph{policy graph}, where nodes describe conversation actions, and edges connect subsequent actions. An \model{} program (Figure \ref{fig:anytod_normal_policy}) is implemented to recommend next actions with respect to this policy graph.

Two baselines are used for comparison: BERT+S \citep{mosig-2020-star-dataset} and SAM \citep{mehri-eskenazi-2021-schema-paradigm}, both of which add a policy graph following module for zero-shot transfer to unseen schema. Note that, though these models were trained on the original \starone{} data, their SaF1 results are directly comparable to \model{} trained on \startwo{} on NAP (SaF1), as these ground truth labels were left untouched. However, \model{} has additional training supervision on AST and DST due to \startwo{}'s new annotations. For a fair comparison with SAM, we also report results on SAM-User, a modified version of SAM trained on \startwo{} that also includes supervised training on user annotations.\footnote{See Section \ref{sec:sam_user_impl} for implementation details.} Note that both BERT+S and SAM are based on BERT-base (110M parameters), comparable to T5 base (220M parameters).

\textbf{Main Result} The primary results for \model\textsc{ base/xxl} are given in Table~\ref{table:star_results}. For conciseness, we shorten \model{} to \modelshort{}. As an ablation, we also report results with \modelshort{}\textsc{-norec}, which removes the policy graph guidance from the \model{} method by recommending no system actions. In the full-shot setting, both \model{} and \textsc{-norec}, along with reported baselines, achieve very high SaF1. This is due to direct supervised training on NAP removing the need for program guidance. However, we see a huge gap between \model{} and \textsc{-norec} in the zero-shot setting; the guidance from the program becomes necessary --- we see 60.6 vs. 55.8 SaF1 at \textsc{base}, and 68.0 vs. 62.3 SaF1 at \textsc{xxl}. Moreover, \model{}\textsc{ xxl} has zero-shot performance comparable to that of full-shot, with 75.4 SaF1 at \textsc{xxl}.

\textbf{Effect of Natural Language Descriptions} As mentioned in Section \ref{sec:starv2_method}, \startwo{} provides new natural language descriptions that better characterize the actions within \starone{}. Our main result \modelshort{}\textsc{ base/xxl} takes advantage of these new descriptions, but to see the impact of these descriptions, we train \modelshort{}\textsc{-tmpl} on the original template utterances. On \textsc{base} we see little difference between descriptions and templates, but a sizeable improvement in using descriptions appears on \textsc{xxl}, with a larger LM that is better at NLU. This evidences that more intuitive natural language descriptions help \model{} understand task semantics better and perform zero-shot transfer.

\textbf{\model{} vs. baselines} To compare against available results on \startwo{}, we compare \modelshort{}\textsc{-tmpl base} against SAM-User. Both results use template responses provided by \starone{}, and additionally trained with the new DST and AST annotations in \startwo{}. However, we see far stronger performance with \model{} than with SAM or SAM-User, due to the flexibility provided by the program execution ability demonstrated by \model{}, and enabled by supervised training on DST and AST. SAM is not suited to use these contextual signals, likely due to no attention between schema elements and conversation and a rigid classification architecture unsuitable for multiple losses.

\textbf{Multitask Training with SGD} To demonstrate further robustness for \model{}, we also report \model{}\textsc{-sgd}, which jointly trains with SGD as a multitask training dataset. SGD includes a large number of tasks, each defined by a schema with highly diverse parameters and actions. The \textsc{-sgd} results in Table \ref{table:star_results} show that at \textsc{base}, SGD multitask training improves both DST ($61.9 \to 66.1$ JGA), AST ($72.1 \to 74.3$ UaF1), and by extension NAP ($60.6 \to 61.3$ SaF1). A similar but smaller improvement is seen on \textsc{xxl}, suggesting that the larger LM may not need more diverse training owing to its better language understanding.

\begin{table}[tb]
\begin{subtable}[h]{\linewidth}
\centering
    \setlength{\tabcolsep}{3.4pt}
    
    \scalebox{0.8}{
    \begin{tabular}{lcHHcHcc}
    \hline 
    \textbf{Model} &\textbf{JGA} &\textbf{bJGA} & \textbf{UaAcc} & \textbf{UaF1} & \textbf{SaAcc} & \textbf{SaF1} & \textbf{BLEU} \\
    \hline
    BERT+S  &- &- &- &- &73.8 &74.9 &- \\
    SAM &- &- &- &- &70.4 &71.5 &- \\
    SAM-User &- &- &- &- &70.4 &71.7 &- \\
    \hline
    \modelshort{}\textsc{-norec base} &81.5& 73.7& 74.4& 83.8& 73.7& 73.3& 72.8 \\
    \modelshort{}\textsc{-tmpl base} &82.9& 74.7& 75.6& 84.6& 71& 70.6& 72.7 \\
    \modelshort{} \textsc{base} &82.4& 74.6& 75.2& 84.1& 71.6& 70.7& 72 \\
    \hline
    \modelshort{}\textsc{-norec xxl} &85.6& 76.7& 78.3& 86.4& 75.7& 75.4& 76.4 \\
    \modelshort{}\textsc{-tmpl xxl} &85.1& 76.3& 72.6& 82.5& 70.7& 71.3& 75.8 \\
    \modelshort{}\textsc{ xxl} &85.7& 76.8& 75.9& 84.7& 73.8& 73.3& 73.5 \\
    \hline
    \end{tabular}
    }
    \caption{Full-shot results on \startwo{}.}
    \label{table:star_full-shot}
    
    \vspace{0.2cm}
    
    \scalebox{0.8}{
    \begin{tabular}{lcHHcHcc}
    \hline
    \textbf{Model} & \textbf{JGA} & \textbf{bJGA} & \textbf{UaAcc} & \textbf{UaF1} & \textbf{SaAcc} & \textbf{SaF1} & \textbf{BLEU} \\
    \hline
    BERT+S  &- &- &- &- &29.7 &32.3 &- \\
    SAM\tablefootnote{Note that this SAM zero-shot domain SaF1 differs from the original 55.7 from \citet{mehri-eskenazi-2021-schema-paradigm}. %
    See Section \ref{sec:appendix_sam_zero_shot_correction} for more details.} &- &- &- &- &49.8 &51.2 &- \\
    SAM-User &- &- &- &- &53.9 &44.4 &- \\
    \hline
    \modelshort{}\textsc{-norec base} &57.8& 55.4& 55.4& 71& 56.1& 55.8& 32.4 \\
    \modelshort{}\textsc{-tmpl base} &62.2& 58.2& 56& 74& 62.5& 61.9& 56 \\
    \modelshort{} \textsc{base} &61.9& 58.2& 56.6& 72.1& 61.6& \textbf{60.6}& 34.3 \\
    \modelshort{}\textsc{-sgd base} &66.1& 62& 59.5& 74.3& 63.5& 61.3& 34.4 \\
    \hline
    \modelshort{}\textsc{-prog base} &61.9& 58.2& 56.6& 72.1& 61.9& 61.0& 34.4 \\
    \modelshort{}\textsc{-prog+sgd base} &66.1& 62& 59.5& 74.3& 64.2& 61.9& 34.6 \\
    \hline
    \modelshort{}\textsc{-norec xxl} &72.7& 67.6& 65.9& 80& 62.3& 62.3& 41.8 \\
    \modelshort{}\textsc{-tmpl xxl} &66.8& 61.3& 58.9& 72.9& 60.9& 60.8& 52.9 \\
    \modelshort{} \textsc{xxl} &74.8& 69.2& 64.6& 79.2& 68& \textbf{68.0}& 44.3 \\
    \modelshort{}\textsc{-sgd xxl} &75.8& 70.4& 67.8& 80.9& 69.3& 68.5& 43.9 \\
    \hline
    \modelshort{}\textsc{-prog xxl} &74.4& 68.9& 64.7& 79.3& 68.5& 68.4& 44.9 \\
    \modelshort{}\textsc{-prog+sgd xxl} &75.7& 70.4& 68.5& 81.4& 70.8& 70.7& 44.2 \\
    \hline
    \end{tabular}
    }
    \caption{Zero-shot domain results on \startwo{.}}
    \label{table:star_zeroshot_domain}

    \scalebox{0.8}{
    \begin{tabular}{lccc}
    \hline
    \textbf{Model} &\textbf{Bank} &\textbf{Trip}  &\textbf{Trivia} \\
    \hline
    \modelshort{}\textsc{ xxl} &54.3& 52.4& 73.8 \\
    \modelshort{}\textsc{-sgd xxl}  &53.1& 51.5& 81.1 \\
    \hline
    \modelshort{}\textsc{-prog xxl} &61& 60.8& 73.7 \\
    \modelshort{}\textsc{-prog+sgd xxl} &65& 62.9& 86.3 \\
    \hline
    \end{tabular}
    }
    \caption{SaF1 on \startwo{} programming tasks..}
    \label{table:star_prog}
\end{subtable}
\caption{Results on \startwo{}. For compactness we show just UaF1 and SaF1 here --- see Section~\ref{sec:appendix_complete_starv2_results} for a complete table. For clarity, we bold SaF1 results for \model{}\textsc{ base/xxl}, our key result.}
\label{table:star_results}
\vspace{-14pt}
\end{table}

\textbf{Complex Program Logic} \startwo{} is also a good testbed for complex zero-shot task adaptation, as it includes some tasks which are more complex than simple policy-graph following, specifically the \texttt{bank}, \texttt{trivia}, and \texttt{trip} domains. For instance, the \texttt{trivia} task requires the agent to ask the user a trivia question and extract their answer. Different system actions must be taken by the agent depending on whether or not the user's answer is correct. This logic is not captured by the provided policy graph alone, requiring more complex logic. \model{} is suitable for this problem, as we need only to construct a program implementing this logic. These programs are shown in Section \ref{sec:anytod_programs_appendix}. 

We report results with these programs in Table \ref{table:star_results} under the \textsc{-prog} name. 
There is a clear win on zero-shot domain SaF1 when averaged over all domains, with a very high 70.7 SaF1 on \textsc{-prog+sgd xxl}, narrowing the gap with the full-shot 75.4 SaF1. When examining the complex tasks tasks individually (Table \ref{table:star_prog}), the win on NAP is even more apparent. The only exception is \modelshort{}\textsc{ xxl}  on \texttt{trivia}, which has little difference with or without the program.
In general however, the guidance provided by this specialized program is necessary for higher-level logic in the dialog policy, since the policy graph does not specify enough information to approach the task in zero-shot.

\subsection{Results on ABCD and SGD} \label{sec:abcd_results}

\begin{table}
    \centering
        \scalebox{0.8}{
    \begin{tabular}{lcccc}
    \hline
    \textbf{Model} & \textbf{JGA} & \textbf{JGA} & \textbf{SaF1}  & \textbf{SaF1} \\
    & seen & unseen & seen & unseen\\
    \hline
    \modelshort{}\textsc{-norec base} & 89.0 & 58.5 & 89.8 & 83.4\\
    \modelshort{}\textsc{ base} & 89.9 & 62.4 & 89.8 & 86.1 \\
    \hline
    \modelshort{}\textsc{-norec xxl} & 94.8 & 80.2 & 92.1 & 87.2 \\
    \modelshort{}\textsc{ xxl} & 94.8 & 82.2 & 91.3 & 88.9 \\
    \hline
    \end{tabular}%
    }
\caption{\model{} JGA, SaF1 on SGD test set.}
\label{table:sgd_policy}
\vspace{-10pt}
\end{table}%

We conduct similar experiments on Action State Tracking (AST) (metric: joint action accuracy or JAA) on ABCD \citep{chen-etal-2021-action} and DST and NAP (metrics: JGA and SaF1 respectively) on SGD \citep{rastogi2020-sgd} datasets.

ABCD contains 10 flows, each describing the business logic for handling a customer request, which are relatively similar to each other. We report full-shot results by training and evaluating on all flows, and zero-shot results where the model is trained on one randomly sampled flow and evaluated on all other nine flows. The SGD test set consists of 21 services, 15 of these not seen during training. The dataset is generated via simulation with a generalized policy graph (shared across all services) encoding dialog act transitions. The per-service policy graphs are then constructed by inserting intents and slots and, as a result, end up similar.

Tables \ref{table:sgd_policy} and \ref{table:abcd_policy} and show \model{} results on SGD and ABCD respectively. For both datasets on both full-shot and zero-shot setups we generally see an improvement on action prediction using policy guidance, achieving state-of-the-art results for ABCD. However, the gain is not as large as \startwo{}, as the task policies are not as diverse. Even without explicit policy guidance, features from different tasks in ABCD/SGD can transfer to each other. Notably, policy guidance helps more on the one-flow setup for ABCD and unseen services for SGD, further establishing the efficacy of policy guidance on unseen setups, even if related.

\begin{table*}[t]
\begin{minipage}{.33\textwidth}
\centering

\resizebox{\textwidth}{!}{%
\setlength{\tabcolsep}{1.4pt}
    \begin{tabular}{lcc}
    \hline
    \textbf{Model} & \textbf{All Flows} & \textbf{One Flow} \\
    \hline
    RoBERTa & 65.8 & - \\
    AST-T5-Small & 87.9 & - \\
    \hline
    \modelshort{}\textsc{-norec base} & 90.5 & 47.4 \\
    \modelshort{}\textsc{ base} & \textbf{90.5} & \textbf{48.9} \\
    \hline
    \modelshort{}\textsc{-norec xxl} & 91.6 & 64.3 \\
    \modelshort{}\textsc{ xxl} & \textbf{91.9} & \textbf{67} \\
    \hline
    \end{tabular}%
}
\caption{JAA on ABCD Action State Tracking (AST) for full-shot (All Flows) and zero-shot transfer (One Flow). The zero-shot JAA is the mean JAA across three experiments.}
\label{table:abcd_policy}
\end{minipage}%
\hspace{5pt}
\begin{minipage}{.22\textwidth}
    \centering
    \setlength{\tabcolsep}{1.4pt}
    \resizebox{\textwidth}{!}{%
    \begin{tabular}{lccc}
    \hline
    \textbf{Model} & \textbf{SaF1} \\
    \hline
    \modelshort{}\textsc{ base} & 60.6 \\
    \modelshort{}\textsc{-0rec base} & 31.3 \\
    \modelshort{}\textsc{-badrec base} & 25.8 \\
    \hline
    \modelshort{}\textsc{ xxl} & 68.0 \\
    \modelshort{}\textsc{-0rec xxl} & 39.3 \\
    \modelshort{}\textsc{-badrec xxl} & 35.0 \\
    \hline
    \end{tabular}
}
    \caption{\startwo{} zero-shot domain SaF1 with \textsc{badrec} and \textsc{0rec}.}
    \label{table:star_badnorec}
\end{minipage}%
\hspace{5pt}
\begin{minipage}{.4\textwidth}%
\includegraphics[width=\columnwidth]{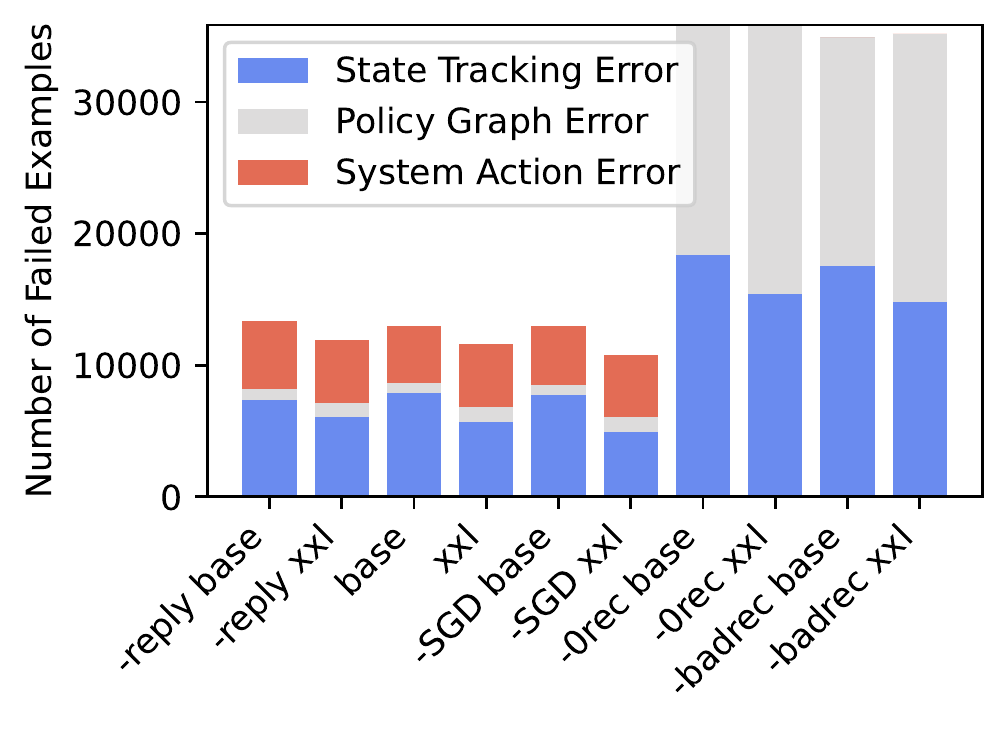}
\vspace{-27pt}
\captionof{figure}{\model{} error analysis on \startwo{} zero-shot domain. %
} 
\label{fig:anytod-error-analysis}

\end{minipage}
\end{table*}

\subsection{Zero-shot Results on MultiWOZ} \label{sec:mwoz_0shot}

\begin{table}[tb]
\begin{subtable}[h]{\linewidth}
\centering
    \setlength{\tabcolsep}{3.4pt}
    
    \scalebox{0.8}{
    \begin{tabular}{lcccc}
    \hline
    \textbf{Model} & \textbf{JGA} & \textbf{Inform} & \textbf{Success} & \textbf{BLEU} \\
    \hline
    SOLOIST & 35.9 & 81.7 & 67.1 & 13.6 \\
    Mars & 35.5 & 88.9 & 78.0 & 19.6 \\
    \hline
    \model{}\textsc{-xxl} & 30.8 & 73.9 & 24.4 & 3.4  \\
    \hline
    \end{tabular}
    }
\end{subtable}
\caption{Results on MultiWOZ end-to-end benchmark. \model{}\textsc{-xxl} is trained on SGD, and evaluated in zero-shot over MultiWOZ. Note we applied templates for response generation, yielding low BLEU in comparison with other models.}
\label{table:mwoz_0shot}
\end{table}

To demonstrate the generalizability of the \model{} system, we demonstrate zero-shot transfer results on the end-to-end MultiWOZ 2.2 \citep{zang-etal-2020-multiwoz} benchmark, a popular dataset for TOD research. In this case, \model{}\textsc{-xxl} is trained on the SGD dataset, and then evaluated on MultiWOZ in zero-shot with a small policy program (Section \ref{sec:mwoz_0shot_policy}). Responses from \model{} were constructed using the template utterance approach from \citet{kale-t2g2}. 
We compare against SOLOIST \citep{Peng2020-soloist} and Mars \citep{Sun2022-mars}, two end-to-end TOD models directly trained on MultiWOZ with supervision. Results are shown in Table \ref{table:mwoz_0shot}, with metrics reported by the MultiWOZ eval script \citep{nekvinda-mwozevalscript}. Although no training examples from MultiWOZ was used at all, \model{} demonstrates strong JGA, Inform, and Success comparable to results that do train on MultiWOZ. Note that since we applied templates for response generation, we do not consider BLEU to be important, as the responses are very different from ground truth labels.

\section{Analysis}\label{sec:analysis}

\subsection{Impact of Policy Guidance}

To see how impactful the recommendations provided by the policy graph are, we reevaluate already finetuned \model{} models on the \startwo{} zero-shot domain setting, but with changes to the program recommendations during eval. First, to see how dependent \model{} is on policy graph guidance, we modify the graph to output no recommendations (denoted as \textsc{0rec}), forcing the model to do NAP only using the conversation, belief state, and action history. Secondly, we modify the graph to output deliberately bad recommendations (denoted as \textsc{badrec}), intended to trick the model into choosing an incorrect system action. This was done by randomly sampling 1-3 system actions other than the ground truth action. %

The major drops in SaF1 for both setups shown in Table \ref{table:star_badnorec} confirm that the model, while able to predict actions without it, does consider the policy guidance heavily. Notably, 75\% and 83\% of correct predictions for \textsc{0rec} and \textsc{badrec} are actions common to all tasks e.g., \texttt{hello} or \texttt{query}.

\begin{table}[t]
\begin{subtable}[h]{\linewidth}
\centering
\setlength{\tabcolsep}{3.4pt}
 \scalebox{0.8}{
 \begin{tabular}{lccc}
 \hline
 \textbf{Model and Corruption Prob.} & \textbf{All Flows} & \textbf{One Flow} \\
 \hline
 \modelshort{}\textsc{ base}, 0 & 90.5 & 48.9 \\
 \modelshort{}\textsc{ base}, 0.4 & 90.1 & 48.4 \\
 \modelshort{}\textsc{ base}, 0.8 & 89.5 & 47.4 \\
 \modelshort{}\textsc{-norec base}, 0 &90.5 & 47.4 \\
 \hline
 \modelshort{}\textsc{ xxl}, 0 & 91.9 & 67 \\
 \modelshort{}\textsc{ xxl}, 0.4 & 91.5 & 66.7 \\
 \modelshort{}\textsc{ xxl}, 0.8 & 91.5 & 65.9 \\
 \modelshort{}\textsc{-norec xxl}, 0 &91.6 & 64.3 \\
 \hline
 \end{tabular}
 }
\end{subtable}
\caption{JAA on ABCD Action State Tracking (AST) with policy corruption. For ``one flow'', the JAA is averaged across three runs with a randomly selected flow for training.}
\label{table:abcd_corrupt}
\vspace{-10pt}
\end{table}

We conduct a similar ``policy corruption'' experiment on ABCD (Table \ref{table:abcd_corrupt}), in which policy graphs for evaluation tasks have a 0\%, 40\%, and 80\% chance of being replaced by graphs from incorrect flows during evaluation. We see a consistent quality drop with increasing probability of corruption for both \textsc{base} and \textsc{xxl}.

\subsection{Error Analysis}

We also analyze \model{} errors on \startwo{}. We classify all incorrect NAPs into three possible error categories: (1) System action error: the program recommends the correct system action, but this was not chosen by the LM, (2) Policy graph error: the predicted belief state and action history are correct, but the program's execution of the policy graph does not recommend the expected system action, and (3) State tracking error: the predicted belief states and action history are incorrect, which leads to incorrect recommendations from the policy graph. Results are shown in Figure \ref{fig:anytod-error-analysis}. In general, we see that the benefit to scaling the LM from \textsc{base} to \textsc{xxl} comes from improvements to state and action tracking, which aligns with better DST and AST results on \textsc{xxl} as in Table \ref{table:star_results}.

\section{Conclusion}

We proposed \model{}, a zero-shot end-to-end TOD system that can be programmed to handle unseen tasks without domain-specific training. \model{} adopts a neuro-symbolic approach, in which a LM performs zero-shot DST and AST with respect to a provided schema, and abstracts both into a sequence of symbols. These symbol sequences are then parsed and passed to a program expressing the task policy, which gets executed to make recommendations for the next agent action(s). Agent designers are free to implement arbitrarily complex business logic within \model{} to determine its policy on unseen tasks or domains. To demonstrate the value of this approach, we show state-of-the-art results on zero-shot TOD benchmarks, such as STAR, ABCD, SGD and MultiWoZ. For further training and benchmarking zero-shot end-to-end TOD systems, we also release the \startwo{} dataset, an improved version of STAR.

\section{Limitations}
\model{} is a data-efficient approach designed to accelerate task-oriented dialog system building. We make use of a relatively large LM in our implementation (T5X, up to 11B parameters) to effectively make structured predictions (dialog states), and further control the language model behavior with symbolic programs designed by the user. While the LM's behavior in our design is properly regulated and we applied templates to formulate responses, one must be responsible in designing the policy program and templates to ensure predictability of the system actions. We do not intend to make use of this system in open-domain, free-form conversation generation scenarios.

\bibliography{anthology,custom}

\appendix

\section{Appendix}
\label{sec:appendix}

\renewcommand{\thefigure}{A.\arabic{figure}}
\setcounter{figure}{0}
\renewcommand{\thetable}{A.\arabic{table}}
\setcounter{table}{0}

\subsection{\model{} Programs} \label{sec:anytod_programs_appendix}

Examples of \model{} program implementations for \startwo{} can be found in Figures \ref{fig:anytod_normal_policy} and \ref{fig:anytod_trivia_bank_policies}.

\subsection{Complete Results on \startwo{}} \label{sec:appendix_complete_starv2_results}

For compactness, Table~\ref{table:star_results} showed just UaF1 and SaF1. We also report user action accuracy (UaAcc) and system action accuracy (SaAcc) in Table \ref{table:star_results_complete}.

\begin{table*}[tb]
\begin{subtable}[h]{\linewidth}
\centering
    \setlength{\tabcolsep}{3.4pt}
    
    \scalebox{1.0}{
    \begin{tabular}{lcHccccc}
    \hline 
    \textbf{Model} &\textbf{JGA} &\textbf{bJGA} & \textbf{UaAcc} & \textbf{UaF1} & \textbf{SaAcc} & \textbf{SaF1} & \textbf{BLEU} \\
    \hline
    BERT+S  &- &- &- &- &73.8 &74.9 &- \\
    SAM &- &- &- &- &70.4 &71.5 &- \\
    SAM-User &- &- &- &- &70.4 &71.7 &- \\
    \hline
    \modelshort{}\textsc{-noguide base} &81.5& 73.7& 74.4& 83.8& 73.7& 73.3& 72.8 \\
    \modelshort{}\textsc{-tmpl base} &82.9& 74.7& 75.6& 84.6& 71& 70.6& 72.7 \\
    \textbf{\modelshort{} \textsc{base}} &82.4& 74.6& 75.2& 84.1& 71.6& 70.7& 72 \\
    \hline
    \modelshort{}\textsc{-noguide xxl} &85.6& 76.7& 78.3& 86.4& 75.7& 75.4& 76.4 \\
    \modelshort{}\textsc{-tmpl xxl} &85.1& 76.3& 72.6& 82.5& 70.7& 71.3& 75.8 \\
    \textbf{\modelshort{}\textsc{ xxl}} &85.7& 76.8& 75.9& 84.7& 73.8& 73.3& 73.5 \\
    \hline
    \end{tabular}
    }
    \caption{Full-shot results on \startwo{}.}
    
    \vspace{0.2cm}
    
    \scalebox{1.0}{
    \begin{tabular}{lcHccccc}
    \hline
    \textbf{Model} & \textbf{JGA} & \textbf{bJGA} & \textbf{UaAcc} & \textbf{UaF1} & \textbf{SaAcc} & \textbf{SaF1} & \textbf{BLEU} \\
    \hline
    BERT+S  &- &- &- &- &29.7 &32.3 &- \\
    SAM &- &- &- &- &49.8 &51.2 &- \\
    SAM-User &- &- &- &- &53.9 &44.4 &- \\
    \hline
    \modelshort{}\textsc{-noguide base} &57.8& 55.4& 55.4& 71& 56.1& 55.8& 32.4 \\
    \modelshort{}\textsc{-tmpl base} &62.2& 58.2& 56& 74& 62.5& 61.9& 56 \\
    \textbf{\modelshort{} \textsc{base}} &61.9& 58.2& 56.6& 72.1& 61.6& 60.6& 34.3 \\
    \modelshort{}\textsc{-sgd base} &66.1& 62& 59.5& 74.3& 63.5& 61.3& 34.4 \\
    \hline
    \modelshort{}\textsc{-prog+reply base} &62.7& 58.6& 55.8& 73.9& 63.1& 62.9& 56.3 \\
    \textbf{\modelshort{}\textsc{-prog base}} &61.9& 58.2& 56.6& 72.1& 61.9& 61.0& 34.4 \\
    \modelshort{}\textsc{-prog+sgd base} &66.1& 62& 59.5& 74.3& 64.2& 61.9& 34.6 \\
    \hline
    \modelshort{}\textsc{-noguide xxl} &72.7& 67.6& 65.9& 80& 62.3& 62.3& 41.8 \\
    \modelshort{}\textsc{-tmpl xxl} &66.8& 61.3& 58.9& 72.9& 60.9& 60.8& 52.9 \\
    \textbf{\modelshort{} \textsc{xxl}} &74.8& 69.2& 64.6& 79.2& 68& 68.0& 44.3 \\
    \modelshort{}\textsc{-sgd xxl} &75.8& 70.4& 67.8& 80.9& 69.3& 68.5& 43.9 \\
    \hline
    \modelshort{}\textsc{-prog+reply xxl} &73.7& 68.2& 61.6& 76.6& 65.7& 66.3& 63.6 \\
    \textbf{\modelshort{}\textsc{-prog xxl}} &74.4& 68.9& 64.7& 79.3& 68.5& 68.4& 44.9 \\
    \modelshort{}\textsc{-prog+sgd xxl} &75.7& 70.4& 68.5& 81.4& 70.8& 70.7& 44.2 \\
    \hline
    \end{tabular}
    }
    \caption{Zero-shot domain results on \startwo{.}}

\end{subtable}
\caption{Complete results on \startwo{}.}
\label{table:star_results_complete}
\end{table*}

\subsection{Corrected SAM Results on Zero-shot Domain} \label{sec:appendix_sam_zero_shot_correction}

During the development of \model{}, we found that the zero-shot domain results reported on SAM in \citet{mehri-eskenazi-2021-schema-paradigm} were incorrect. An annotation issue within the STAR dataset set marked some conversations as having an invalid domain; due to how SAM was implemented, these conversations would always be included in the training dataset, even if they were in the evaluation domain. For instance, dialog ID \texttt{102} is marked as a \texttt{null} domain in the original \starone{} dataset. Retraining SAM with this issue fixed caused a drop in SaF1, from 55.7 to 51.2. We fix these annotation errors in the \startwo{} dataset.

\subsection{Calculating \startwo{} Metrics} \label{sec:appendix_starv2_metrics_calc}

Details in calculating metrics on \startwo{} are as follows. For DST, JGA is calculated with an exact match on belief state parameters and values. For AST, we only consider the quality of the most recent turn within the action history prediction. This is always a user turn, which may have multiple user actions active. This may be considered a multilabel classification problem. Then, we calculate UaAcc through exact set match on the predicted user actions at the current turn, as well as weighted multilabel F1 on the predicted user actions. Both SaAcc and SaF1 are calculated as described in \citet{mosig-2020-star-dataset}.

\subsection{Implementation of Sam-User} \label{sec:sam_user_impl}

To implement supervised training of AST on SAM, we modify the methodology described in \citet{mehri-eskenazi-2021-schema-paradigm}, which embeds both conversation and schema elements to produce an attention vector $p$. Here, $p_i$ gives the attention weight between the conversation and the $i$-th user action of the policy graph. This is then interpreted to be a proxy for probability, and converted to a probability for NAP on all system actions $a$ according to the policy graph edges:
\begin{gather*}
    g(i,a) = \begin{cases}
        p_i, & \text{if } \textbf{action}(\textbf{next}(u_i)) = a\\
        0, & \text{otherwise}
    \end{cases}\\
    P(a) = \sum_{i \leq |S|} g(i,a)
\end{gather*}
Here, $\textbf{action}(\textbf{next}(u_i))$ gives the next system action of the user action $u_i$ according to the policy graph. Note that $p_i$ is an attention weight that is interepted to be the probability of user action $u_i$ being active at the current turn; however, no supervised training was done with ground truth user action labels. Then, to implement supervised training on these user actions, we train $p_i$ to be actual probabilities, and apply a sigmoid on $p_i$ to form a user action prediction head. Note that this is a multilabel binary prediction. We then calculate a binary cross-entropy loss on this head.

\subsection{MultiWOZ Zero-Shot Policy Program} \label{sec:mwoz_0shot_policy}

Figure \ref{fig:mwoz_policy} contains the \model{} policy program used when evaluating over MultiWOZ. This policy program was handcrafted, and provides a simplified conversation flow.

\begin{figure*}[ht]
\begin{lstlisting}[language=Python]
def multiwoz_policy(active_domain, belief_state, act_hist):
  rec = []
  last_useracts = act_hist[-1]

  # We define a new action within the MultiWOZ schema that tracks whether
  # the user wants to book a provided entity.
  # Since this is zero-shot we don't train on this action at all, just provide
  # a natural language description "user is saying they wants to book this hotel"
  user_wants_to_book = any(act == 'user-wants-to-book' for act, _ in last_useracts)

  if user_wants_to_book:
    rec.append(('book', None))
    # Inform the name of what we're booking for the user
    if active_domain in ['restaurant', 'hotel', 'attraction']:
      rec.append(('inform', f'{active_domain}-name'))
    elif active_domain == 'train':
      rec.append(('inform', f'train-trainid'))
    # Ask the user if they need anything else
    rec.append(('reqmore', None))
  else:
    # We're still trying to find an entity for the user
    # Recommend / select entities
    if active_domain in ['restaurant', 'hotel', 'attraction']:
      rec.append(('inform', f'{active_domain}-name'))
    elif active_domain == 'train':
      rec.append(('inform', f'train-trainid'))
    rec.append(('recommend', None))
    rec.append(('select', None))
    rec.append(('booking-inform', None))

  for act, slot in last_useracts:
    if act == 'inform':
      # We often repeat back info the user has given us in next turn
      rec.append(('inform', slot))
    # If the user is requesting a slot, provide the value
    if act == 'request':
      rec.append(('inform', slot))
    # If the user is thanking us, say you're welcome / bye / anything else?
    if act == 'thank':
      rec.append(('welcome', None))
      rec.append(('bye', None))
      rec.append(('reqmore', None))

  return set(rec)
\end{lstlisting}
\caption{The \model{} program implementation for the zero-shot policy program.}
\label{fig:mwoz_policy}
\end{figure*}

\begin{figure*}[ht]
\begin{lstlisting}[language=Python]
USER_CUSTOM_LABEL = 'user_custom'
OUT_OF_SCOPE_LABEL = 'out_of_scope'

def anytod_star_policy_program(
    belief_state: dict[str, str], act_hist: list[list[str]], api: Json,
    graph: Json, convo_hist: list[str], primary_item: Json):
  # a list of next action prdictions to recommend to the lm
  next_act_recs = []
  # get the "bye" actions for both user and system
  user_bye_act = _user_bye_act(graph)
  sys_bye_act = _sys_bye_act(graph)

  # dict of param -> action user would take to inform this param
  slot_actions = graph['slot_actions']
  # generate a list of all user informing acts
  inform_user_acts = set()
  for _, user_acts in slot_actions.items():
    inform_user_acts.add(user_acts[0])

  if act_hist:
    # iterate through last turn's active user actions, result of AST
    for last_useract in act_hist[-1]:
      # some transitions are common to all star graphs, but not explicit
      # if user is performing something out-of-scope, return out_of_scope
      if last_useract == USER_CUSTOM_LABEL:
        next_act_recs.append(OUT_OF_SCOPE_LABEL)
      # if user is saying bye, agent can say bye
      if last_useract == user_bye_act:
        next_act_recs.append(sys_bye_act)

      # if the user is performing an action that isn't informing a param,
      # look it up in the policy graph
      if last_useract not in inform_useracts and last_useract in graph['graph']:
        next_act_recs.append(graph['graph'][last_useract])
  # if the agent can do the anything_else action, it can also say bye
  if 'anything_else' in next_act_recs:
    next_act_recs.append(bye_act)

  # if all required params are provided, we can query api
  query_label = 'query' if 'query' in graph['replies'] else 'query_check'
  if all(p.name in belief_state for p in api.params if p.required):
    next_act_recs.append(query_label)

  # param name -> api param json
  api_params_by_name = {}
  for param in api['input']:
    if param['Name'] != 'RequestType':
      api_params_by_name[param['Name']] = param
  # if a param is not known, we can request it from the user
  for slot in graph['slot_actions']:
    p = api_params_by_name[slot]
    if p.name not in belief_state:
      ask_sysact = slot_actions[p.name][0]
      next_act_recs.append(ask_sysact)

  return next_act_recs
\end{lstlisting}
\caption{The \model{} program implementation for a given \starone{} policy graph.}
\label{fig:anytod_normal_policy}
\end{figure*}

\begin{figure*}[ht]
\begin{lstlisting}[language=Python]
def anytod_star_trivia_policy(
    belief_state: dict[str, str], act_hist: list[list[str]], api: Json,
    graph: Json, convo_hist: list[str], primary_item: Json):
  if act_hist and len(convo_hist) >= 2:
    for last_useract in act_hist[-1]:
      # if the user is answering a question
      if last_useract == 'user_trivia_answer':
        # check that the correct trivia answer is in the user's utterance
        answer = primary_item.get('Answer', None)
        if answer:
          last_user_utt = convo_hist[-2]
          if answer.lower() in last_user_utt.lower():
            return ['trivia_inform_answer_correct_ask_next']
          else:
            return ['trivia_inform_answer_incorrect_ask_next']
  return normal_policy(belief_state, act_hist, api, graph, convo_hist,
                       primary_item)


def anytod_star_bank_policy(
    belief_state: dict[str, str], act_hist: list[list[str]], api: Json,
    graph: Json, convo_hist: list[str], primary_item: Json):
  # next_act_recs should be populated already by graph following
  # same as normal_policy() ...

  # params required for authenticating first and second way
  first_auth_slots = ['FullName', 'AccountNumber', 'PIN']
  second_auth_slots = [
      'FullName', 'DateOfBirth', 'SecurityAnswer1', 'SecurityAnswer2'
  ]
  # if either params are satisfied we can query api
  if (all(slot in bs for slot in first_auth_slots) or
      all(slot in bs for slot in second_auth_slots)):
    next_action_recs.append('query')

  # get all seen user acts
  seen_useracts = set()
  for turn, turn_acts in enumerate(act_hist):
    if turn %
      seen_useracts.update(turn_acts)
  forgot_acts = ['user_bank_forgot_account_number', 'user_bank_forgot_pin']
  # if the user has forgotten anything from first auth, follow second auth
  is_second_auth = any(fa in seen_useracts for fa in forgot_acts)
  slots = second_auth_slots if is_second_auth else first_auth_slots
  if graph['task'] == 'bank_fraud_report':
    slots.append('FraudReport')
  # request slots depending on 1st/2nd auth if not known
  for slot in slots:
    if slot not in belief_state:
      next_act_recs.append(graph['slot_actions'][slot][0])

  return next_act_recs
\end{lstlisting}

\caption{The \model{} program implementation specialized for \texttt{bank} and \texttt{trivia} domains.}
\label{fig:anytod_trivia_bank_policies}
\end{figure*}

\end{document}